%% file: main.tex
\documentclass[letterpaper]{article} 
\usepackage{aaai23}  
\usepackage{times}  
\usepackage{helvet}  
\usepackage{courier}  
\usepackage[hyphens]{url}  
\usepackage{graphicx} 
\usepackage{natbib}  
\usepackage{caption} 
\usepackage{algorithm}
\usepackage{algorithmic}
\usepackage{multirow}
\usepackage{comment} 
\usepackage{subcaption}
\usepackage{breqn}

\usepackage{newfloat}
\usepackage{listings}
\DeclareCaptionStyle{ruled}{labelfont=normalfont,labelsep=colon,strut=off} 
\floatstyle{ruled}
\newfloat{listing}{tb}{lst}{}
\floatname{listing}{Listing}
\begin{document}
\title{Can Bad Teaching Induce Forgetting? Unlearning in Deep Networks Using an Incompetent Teacher}
\author{
    Vikram S Chundawat\equalcontrib\textsuperscript{\rm 1}, Ayush K Tarun\equalcontrib \textsuperscript{\rm 1}, Murari Mandal\textsuperscript{\rm 2}\thanks{Work performed while at the School of Computing, National University of Singapore}\thanks{Corresponding author}, Mohan Kankanhalli \textsuperscript{\rm 3}
}
\affiliations{
    \textsuperscript{\rm 1}Mavvex Labs, India\\
    \textsuperscript{\rm 2}School of Computer Engineering, Kalinga Institute of Industrial Technology Bhubaneswar\\
    \textsuperscript{\rm 3}School of Computing, National University of Singapore\\
    \{vikram2000b, ayushtarun210\}@gmail.com, murari.mandalfcs@kiit.ac.in, mohan@comp.nus.edu.sg
}

\maketitle
\input{sec/0_abstract}
\input{sec/1_introduction}
\input{sec/2_related}
\input{sec/3_method}
\input{sec/4_results}

\input{sec/5_conclusions}

{
    \small
    \bibliography{main}
}

\input{sec/X_supplementary}

\end{document}

%% file: sec/0_abstract.tex
\begin{abstract}
Machine unlearning has become an important area of research due to an increasing need for machine learning (ML) applications to comply with the emerging data privacy regulations. It facilitates the provision for removal of certain set or class of data from an already trained ML model without requiring retraining from scratch. Recently, several efforts have been put in to make unlearning to be effective and efficient. We propose a novel machine unlearning method by exploring the utility of competent and incompetent teachers in a student-teacher framework to induce forgetfulness. The knowledge from the competent and incompetent teachers is selectively transferred to the student to obtain a model that doesn't contain any information about the forget data. We experimentally show that this method generalizes well, is fast and effective. Furthermore, we introduce the \textit{zero retrain forgetting (ZRF) metric} to evaluate any unlearning method. Unlike the existing unlearning metrics, the ZRF score does not depend on the availability of the expensive retrained model. This makes it useful for analysis of the unlearned model after deployment as well. We present results of experiments conducted for random subset forgetting and class forgetting on various deep networks and across different application domains.~Source code is at:~\url{https://github.com/vikram2000b/bad-teaching- unlearning}
\end{abstract}


%% file: sec/1_introduction.tex
\section{Introduction} \label{sec:intro}
Machine learning (ML) models are being widely deployed for various applications across different organizations. These models are often trained with large-scale user data. Modern data regulatory frameworks such as European Union GDPR~\cite{voigt2017eu}, and California Consumer Privacy Act (CCPA)~\cite{goldman2020introduction} provide for citizens the {\em right to be forgotten}. It mandates deletion-upon-request of user data. The regulations also require that user consent must be obtained prior to data collection. This consent for the use of an individual's data in these ML models may be withdrawn at any point of time. Thus, a request for data deletion can be made to the ML model owner. The owner company (of the ML model) is legally obligated to remove the models/algorithms derived from using that particular data. As the ML models usually memorize the training samples~\cite{feldman2020does,carlini2019secret}, the company either needs to retrain the model from scratch by excluding the requested data or somehow erase the user's information completely from the ML model parameters. The algorithms supporting such information removal are known as~\textit{machine unlearning} methods. Machine unlearning also offers a framework to prove data removal from the updated ML model.\par

The unlearning methods can be practically applied in the following ways: (i) forgetting single-class or multiple classes of data~\cite{tarun2021fast}, (ii) forgetting a cohort of data from a single class~\cite{golatkar2020eternal,golatkar2020forgetting}, (iii) forgetting a random subset of data from multiple classes~\cite{golatkar2021mixed}. In this paper, we investigate the utility of teacher-student framework with knowledge distillation to develop a robust unlearning method that can support all the three modes, i.e. single/multiple class-level, sub-class level and random subset-level unlearning. Another important question we raise is \textit{how well the unlearned model has generalized the forgetting?} Recent studies suggest that the unlearning methods may lead to privacy leakage in the models~\cite{chen2021machine}. Therefore, it is important to validate whether the unlearned models are susceptible to privacy attacks such as membership inference attacks. Moreover, the trade-off between the amount of unlearning and privacy exposure also should be investigated for better decision-making on the part of the model owner. We propose a new metric to evaluate the generalization ability of the unlearning method.\par

The existing unlearning methods for deep networks put several constraints over the training procedure. For example,~\cite{golatkar2021mixed} train an additional mixed-linear model along with the actual model which is used in their unlearning method. Similarly,~\cite{golatkar2020eternal,golatkar2020forgetting} strictly require SGD to be used in optimization during model training. These restrictions and the need for other prior information make these methods less practical for real-world applications. We present a method that does not require any prior information about the training procedure. We do not train any extra models to assist in the unlearning. Furthermore, we aim to keep the unlearning process efficient and fast in comparison to the high computational costs of the existing methods.

We make the following key contributions:
\begin{enumerate}
     \item We present a teacher-student framework, consisting of competent and incompetent teachers. The selective knowledge transfer to the student results in the unlearned model. The method works for both single-class and multiple class unlearning. It also works effectively for multiple class random-subset forgetting.  
    
    \item We propose a new retrained model-free evaluation metric called zero retrain forgetting (ZRF) metric  to robustly evaluate the unlearning method. This also helps in assessing the generalization in the unlearned model on the forget data.  
    
    \item Our method works on different modalities of deep networks such as CNN, Vision transformers, and LSTM. Unlike the existing methods, our method doesn't put any constraints over the training procedure. We also demonstrate the wide applicability of our method by conducting experiments in different domains of multimedia applications including image classification, human activity recognition, and epileptic seizure detection.
\end{enumerate}

%% file: sec/2_related.tex
\section{Related Work}
\label{sec:related}

\textbf{Machine Unlearning.} Bourtoule et al.~\cite{bourtoule2021machine} proposed to partition the training dataset into non-overlapping shards and create multiple models for the disjoint sets. They store the weakly learned models to deal with multiple data removal requests.~Ginart et al.~\cite{ginart2019making} adopted the definition of differential privacy to introduce the probabilistic notion of unlearning. It expects high similarity between the output distributions of the unlearned model and the retrained model without using the deletion data. Several subsequent works~\cite{mirzasoleiman2017deletion,izzo2021approximate,ullah2021machine} follow this approach in presenting theoretical guarantees in their respective problem settings. We also follow this definition of unlearning in our work.~Guo et al.~\cite{guo2020certified} give a certified data removal framework to enable data deletion in linear and logistic regression.~Neel et al.~\cite{neel2021descent} apply gradient descent to achieve unlearning in convex models.
The difference between differential privacy and machine unlearning is studied in~\cite{sekhari2021remember}. Unlearning in random forests~\cite{brophy2021machine} and Bayesian setting~\cite{nguyen2020variational} are also studied. These methods are designed specifically for convex problems and are unlikely to work in deep learning models. Our work is aimed at performing unlearning in deep networks.\par  

\textbf{Unlearning in Deep Networks.}
Golatkar et al.~\cite{golatkar2020eternal} presented one of the early works in deep machine unlearning. They introduced a scrubbing method to remove the information from the network weights. The method impose a condition of SGD based optimization during training. The subsequent work~\cite{golatkar2020forgetting} proposed a neural tangent kernel (NTK) based method to approximate the training process. The additional approximated model is used to estimate the network weights for the unlearned model.~\cite{golatkar2021mixed} train a mixed-linear model along with the original model. The linearized model is specific to different deep networks and requires fine-tuning to work properly. Moreover, all these methods suffer from high computational costs, constraints on the training process, and limitations of the approximation methods. Tarun et al.~\cite{tarun2021fast} proposed an efficient class-level machine unlearning method. However, it does not support random subset forgetting. In our work, we do not need to train any additional model to support unlearning. Our method does not demand the use of any specific optimization technique during training or any other prior information about the training process.~\cite{chundawat2022zero,graves2021amnesiac,tarun2022deep} are some other notable works.\par

%% file: sec/3_method.tex
\section{Preliminaries}
Let the complete (multimedia) dataset be $D_c = \{(x_i, y_i)\}_{i = 1}^n$ with $n$ number of samples, where $x_i$ is the $i^{th}$ sample, and $y_i$ is the corresponding class label. The set of samples to forget is denoted as $D_f$. In class-level unlearning, $D_f$ corresponds to all the data samples present in a single or multiple classes. In random-subset unlearning, $D_f$ may either consist of a random subset of data samples from a single class or multiple classes. The information exclusive to these data points need to be removed from the model. The set of remaining samples to be retained is denoted by $D_r$. The information about these samples are to be kept unchanged in the model. $D_f$ and $D_r$ together represent the whole training set and are mutually exclusive, i.e. $D_r\cup D_f = D_c$ and $D_r \cap D_f = \phi$. Each data point is assigned an unlearning label, $l_u$, which is $1$ if the sample belongs to $D_f$ and $0$ if it belongs to $D_r$. The subset used for unlearning is $ \{(x_i, l_{u_{i}})\}_{i = 1}^p$, $p$ is total number of samples, and $l_{u_{i}}$ is unlearning label corresponding to each sample $x_i$.\par

The model trained from scratch without observing the forget samples is called the \textit{retrained model} or the \textit{gold model} in this paper. In the proposed teacher-student framework, the \textit{competent teacher} is the fully trained model or the original model. The competent teacher has observed and learned from the complete data $D_{c}$. Let $T_s(x;\theta)$ denote the competent/smart teacher with parameters $\theta$. It takes $x$ as input and outputs the probabilities $t_{s}$. The \textit{incompetent teacher} is a randomly initialized model. Let $T_d(x;\phi)$ be the incompetent/dumb teacher with parameters $\phi$ and output probabilities $t_{d}$. The student $S(x;\theta)$ is a model initialized with parameters $\theta$ i.e., the same as the competent teacher. It returns the output probabilities $s$. It is to be noted that the student is initialized with all the information present in the original model ($\theta$). The incompetent teacher is used to remove the requested information (about the forget data $D_f$) from this model. The Kullback-Leibler (KL) divergence~\cite{kullback1951information} is used as a measure of similarity between two probability distributions. For two distributions $p(x)$ and $q(x)$, the KL-divergence is defined by

\section{Proposed Method}
\subsection{Unlearning with Competent/Incompetent Teachers}
\begin{figure}[]
    \centering
    \includegraphics[width=0.8\linewidth]{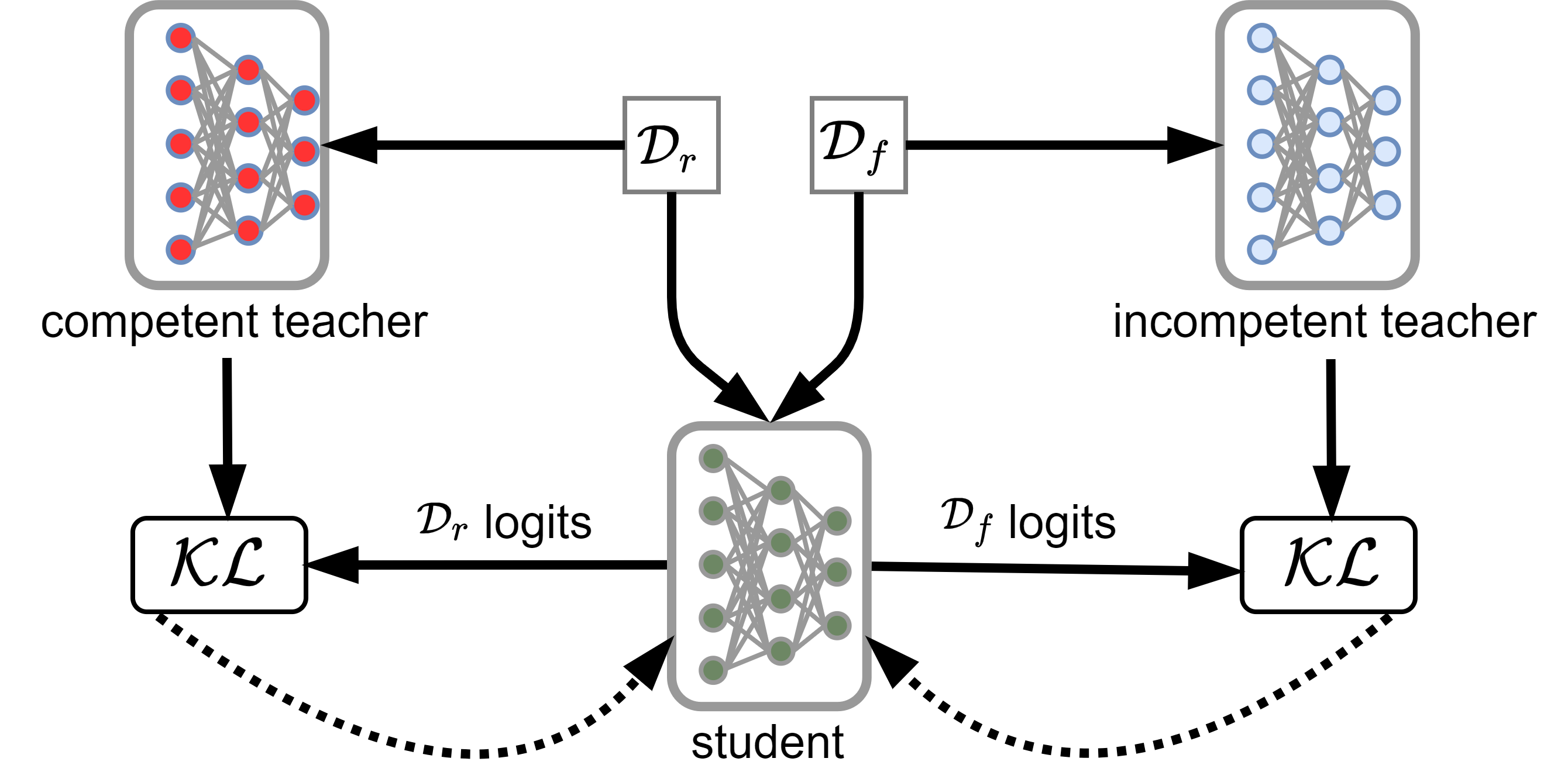}
    \caption{The proposed competent and incompetent teachers based framework for unlearning}
    \vspace{-1.5em}
    \label{fig:teacher-student}
\end{figure}

We aim to remove the information about the requested data-points by using two teachers (competent and incompetent) and one student. The student is initialized with knowledge about the complete data i.e., the parameters of the fully trained model. The idea is to \textit{selectively remove the information} about the forget samples from this model. At the same time, the information pertaining to the retain set should not to be disturbed. Thus, the unlearning objective is to remove the information about $D_f$ while retaining the information about $D_r$. We achieve this by using a pair of (competent/smart ($T_s$) and incompetent/dumb ($T_d$)) teachers to manipulate the student ($S$) as depicted in Figure~\ref{fig:teacher-student}. The bad knowledge about $D_f$ from the incompetent teacher $T_d$ is passed on to the student which helps the student to forget $D_f$ samples. Such an approach consequently induces random knowledge about the forget set in the student instead of completely making their prediction accuracy zero. This serves as a protection against the risk of information exposure about the samples to forget. The bad (random) inputs from $T_d$ may invariably corrupt some of the information about the retain set $D_r$ in the student. Therefore, we selectively borrow correct knowledge related to $D_r$ from the competent teacher $T_{s}$ as well. In this manner, both the incompetent and competent teachers help the student forget and retain the corresponding information, respectively.\par

For a student $S$, incompetent/dumb teacher $T_d$, and competent/smart teacher $T_s$, we define the KL-Divergence between $T_d$ and $S$ in Eq.~\ref{eq1}.

\begin{equation}
\label{eq1}
    \mathcal{KL}(T_d(x)||S(x)) = \sum_i{t_d^{(i)}log(t_d^{(i)}/s^{(i)})}
\end{equation}
where $i$ corresponds to the data class. Similarly, the KL-Divergence between the fully trained competent teacher $T_s$ and student $S$ is given in Eq.~\ref{eq2}.
\begin{equation}
\label{eq2}
    \mathcal{KL}(T_s(x)||S(x)) = \sum_i{t_s^{(i)}log(t_s^{(i)}/s^{(i)})}
\end{equation}

The unlearning objective can be formulated as in Eq.~\ref{eq3}.
\begin{dmath}
\label{eq3}
    L(x, l_u) = (1 - l_u)*\mathcal{KL}(T_s(x)||S(x)) + l_u*(\mathcal{KL}(T_d(x)||S(x)))
\end{dmath}
where $l_u$ is the unlearning label and $x$ is a data sample. The data samples used by the proposed unlearning method consists of all the samples from $D_f$ and a small subset of samples of $D_r$.
The student is then trained to optimize the loss function $L$ for all these samples. The intuition behind optimizing over $L$ is that we selectively transfer bad knowledge about forget data $D_f$ from $T_d$  by minimizing KL-Divergence between $S$ and $T_{d}$ and the accurate knowledge corresponding to $D_r$ is fed from $T_s$ by minimizing KL-Divergence between $S$ and $T_{s}$. The student learns to mimic $T_d$ for $D_f$, thus removing information exclusively pertaining to those samples while retaining all the generic information which can be obtained by other samples of same class.

\subsection{Zero Retrain Forgetting Metric}
\label{sec_zrf}
The effectiveness of an unlearning method is evaluated employing several metrics in the literature. Some frequently used metrics are `accuracy on forget set and retain set'~\cite{golatkar2020eternal,tarun2021fast,golatkar2021mixed,chundawat2022zero}, relearn time~\cite{tarun2021fast}, membership inference attacks~\cite{golatkar2021mixed,graves2021amnesiac}, activation distance~\cite{golatkar2020eternal,golatkar2021mixed}, Anamnesis Index~\cite{chundawat2022zero}, and layer-wise distance~\cite{tarun2021fast}. Excluding the forget and retain set accuracy, all of the remaining metrics in the literature \textit{require a retrained model} i.e., training a model from scratch without using the forget set. These metrics can only be interpreted with reference to such a retrained model. Such dependency on the retrained model for unlearning evaluation would lead to higher time and computational costs. Simply measuring the performance on $D_f$ and $D_r$ does not reveal whether the information is actually removed from the network weights. Thus it is not a comprehensive measure of unlearning.


We propose a novel `Zero Retrain Forgetting Metric' (ZRF) to enable evaluation of unlearning methods \textit{free from dependence on the retrained model}. It measures the randomness in the model's prediction by comparing them with the incompetent teacher $T_d$. We calculate the Jensen–Shannon (JS) divergence\cite{lin1991divergence} between an unlearned model $M$ and the incompetent teacher $T_d$ as below. 

\begin{dmath}
    \mathcal{JS}(M(x), T_d(x)) = 0.5 * \mathcal{KL}(M(x) \| m) + 0.5 * \mathcal{KL}(T_d(x) \| m)
\end{dmath}
where $m = \frac{M(x) + T_d(x)}{2}$. The ZRF metric is defined as
\begin{equation}
    \mathcal{ZRF} = 1 -  \frac{1}{n_f}\sum_{i=0}^{n_f}\mathcal{JS}(M(x_i), T_d(x_i))
\end{equation}
where $x_i$ is $i^{th}$ sample from $D_f$ with a total of $n_f$ samples. The ZRF compares the output distribution for the forget set in the unlearned model with the output of a randomly initialized model, which is our incompetent teacher in most of the cases. The ZRF score lies between 0 and 1. The score will be close to 1 if the model behaviour is completely random  for the forget samples and it will be close to 0 if the model shows some specific pattern.\par

\textbf{What is an ideal ZRF score?} Suppose there is a class \textit{aeroplanes} that contains images of \textit{Boeing aircraft} along with other aircraft models in the training set. If we unlearn \textit{Boeing aircraft}, we don't expect the model to now classify them as \textit{animals}, \textit{vegetables} or any other totally unrelated class. We still expect most of these unlearned images to be classified as aeroplanes. This comes from the intuition that the model must have been designed and trained with generalization in mind. An unlearning method that makes the performance much worse than the generalization error for \textit{aeroplanes} is not actually unlearning. It is just teaching the model to be consistently incorrect when it sees a \textit{Boeing aeroplane}. The ZRF score will be 0 when the model almost always classifies a \textit{Boeing aircraft} as an \textit{animal} or some other totally different class. The ZRF will be 1 if the model always classifies all classes with same random probability for \textit{Boeing aircraft}. Both of these ($\sim0$ or $\sim1$) are not the desirable outcomes. We expect the unlearned model to have a generalization performance similar to that of a model trained without the \textit{Boeing aircraft}. It will have some random predicted logits since the \textit{Boeing aircraft} class was not overfitted during training.

An ideal value of ZRF score depends on the model, dataset and the forget set. Ideally, the optimal ZRF value is what a model trained without the forget set would have. But in practical scenarios we do not have access to the retrained model. So, a good proxy for the ideal ZRF value could be the ZRF value obtained on a test set. The test set by definition is a set about which the model has never learned anything specifically. It is equivalent to saying, a \textit{set} that the model has unlearned perfectly. 



%% file: sec/4_results.tex
\begin{table*}[t]
\centering
\begin{tabular}{c|c|c|c|ccc|ccc|c|ccc}
\hline
\multirow{2}{*}{} & Super & Sub &  
\multicolumn{4}{c|}{Accuracy ($D_f$ $\downarrow$, $D_r$ $\uparrow$)} & \multicolumn{3}{c|}{ZRF} & JS- & \multicolumn{3}{c}{Mem. Attack Prob}\\
\cline{8-10}
\cline{4-7}
\cline{12-14}
 &  {Class} & {Class} & {Acc.} & Orig. & Retrain & \textbf{\textbf{Our}}  & Orig. & Retrain & \textbf{\textbf{Our}} & Div & Orig. & Retrain & \textbf{Our}\\
\hline
\multirow{10}{*}{\rotatebox[origin=c]{90}{ResNet18}} & \multirow{2}{*}{Veh2} & \multirow{2}{*}{Rocket} & $D_r$  & 85.78 & 85.79 & 85.05$\pm$0.61 & \multirow{2}{*}{0.87} & \multirow{2}{*}{0.93} & \multirow{2}{*}{0.99} & \multirow{2}{*}{0.04} & \multirow{2}{*}{0.98} & \multirow{2}{*}{0.52} & \multirow{2}{*}{0.00}\\
{} & {} & {} & $D_f$  & 82 & 3 & 2$\pm$0.40 & {} & {} & {} & {} & {} & {} & {}\\
\cline{2-14}
 & \multirow{2}{*}{Veg} & \multirow{2}{*}{MR} & $D_r$  & 85.82 & 85.38 & 84.79$\pm$0.51 & \multirow{2}{*}{0.88} & \multirow{2}{*}{0.93} & \multirow{2}{*}{0.99} & \multirow{2}{*}{0.04} & \multirow{2}{*}{0.99} & \multirow{2}{*}{0.41} & \multirow{2}{*}{0.00}\\
{} & {} & {} & $D_f$  & 78 & 4 & 1$\pm$0.36 & {} & {} & {} & {} & {} & {} & {}\\
\cline{2-14}
 & \multirow{2}{*}{People} & \multirow{2}{*}{Baby} & $D_r$  & 85.67 & 85.73 & 85.18$\pm$0.54 & \multirow{2}{*}{0.84} & \multirow{2}{*}{0.87} & \multirow{2}{*}{0.98} & \multirow{2}{*}{0.05} & \multirow{2}{*}{1.0} & \multirow{2}{*}{0.84} & \multirow{2}{*}{0.58}\\
 &  &  & $D_f$  & 93 & 82 & 77$\pm$0.34 &  &  &  &  &  &  & \\
\cline{2-14}
 & \multirow{2}{*}{ED} & \multirow{2}{*}{Lamp} & $D_r$  & 85.83 & 86.28 & 84.74$\pm$0.32 & \multirow{2}{*}{0.88} & \multirow{2}{*}{0.94} & \multirow{2}{*}{0.98} & \multirow{2}{*}{0.03} & \multirow{2}{*}{0.98} & \multirow{2}{*}{0.43} & \multirow{2}{*}{0.01}\\
{} & {} & {} & $D_f$  & 77 & 14 & 5$\pm$0.29 &  &  &  &  &  &  & \\
\cline{2-14}
 & \multirow{2}{*}{NS} & \multirow{2}{*}{Sea} & $D_r$  & 85.63 & 85.46 & 84.58$\pm$0.22 & \multirow{2}{*}{0.84} & \multirow{2}{*}{0.87} & \multirow{2}{*}{0.98} & \multirow{2}{*}{0.07} & \multirow{2}{*}{0.99} & \multirow{2}{*}{0.88} & \multirow{2}{*}{0.42}\\
{} & {} & {} & $D_f$  & 97 & 83 & 84$\pm$0.68 & {} & {} & {} & {} & {} & {} & {}\\
\hline

\multirow{10}{*}{\rotatebox[origin=c]{90}{ViT}} & \multirow{2}{*}{Veh2} & \multirow{2}{*}{Rocket} & $D_r$  & 94.89 & 95.35 & 94.84$\pm$0.71 & \multirow{2}{*}{0.91} & \multirow{2}{*}{0.96} & \multirow{2}{*}{0.99} & \multirow{2}{*}{0.03} & \multirow{2}{*}{0.99} & \multirow{2}{*}{0.47} & \multirow{2}{*}{0.01}\\
{} & {} & {} & $D_f$  & 98 & 9 & 17$\pm$0.2 & {} & {} & {} & {} & {} & {} & {}\\
\cline{2-14}
 & \multirow{2}{*}{Veg} & \multirow{2}{*}{MR} & $D_r$  & 94.94 & 94.8 & 94.59$\pm$0.65 & \multirow{2}{*}{0.93} & \multirow{2}{*}{0.98} & \multirow{2}{*}{0.99} & \multirow{2}{*}{0.02} & \multirow{2}{*}{0.99} & \multirow{2}{*}{0.25} & \multirow{2}{*}{0.01}\\
{} & {} & {} & $D_f$  & 93 & 4 & 17$\pm$0.54 & {} & {} & {} & {} & {} & {} & {}\\
\cline{2-14}
 & \multirow{2}{*}{People} & \multirow{2}{*}{Baby} & $D_r$  & 94.91 & 95.26 & 94.45$\pm$0.69 & \multirow{2}{*}{0.90} & \multirow{2}{*}{0.92} & \multirow{2}{*}{0.99} & \multirow{2}{*}{0.06} & \multirow{2}{*}{1.0} & \multirow{2}{*}{0.91} & \multirow{2}{*}{0.14}\\
{} & {} & {} & $D_f$  & 96 & 92 & 77$\pm$0.33 & {} & {} & {} & {} & {} & {} & {}\\
\cline{2-14}
 & \multirow{2}{*}{ED} & \multirow{2}{*}{Lamp} & $D_r$  & 94.93 & 94.86 & 94.990.85 & \multirow{2}{*}{0.91} & \multirow{2}{*}{0.96} & \multirow{2}{*}{0.99} & \multirow{2}{*}{0.03} & \multirow{2}{*}{1.0} & \multirow{2}{*}{0.57} & \multirow{2}{*}{0.02}\\
{} & {} & {} & $D_f$  & 94 & 13 & 21$\pm$0.29 & {} & {} & {} & {} & {} & {} & {}\\
\cline{2-14}
 & \multirow{2}{*}{NS} & \multirow{2}{*}{Sea} & $D_r$  & 94.91 & 94.93 & 94.31$\pm$0.46 & \multirow{2}{*}{0.90} & \multirow{2}{*}{0.92} & \multirow{2}{*}{0.99} & \multirow{2}{*}{0.06} & \multirow{2}{*}{1.0} & \multirow{2}{*}{0.97} & \multirow{2}{*}{0.12}\\
{} & {} & {} & $D_f$  & 96 & 85 & 79$\pm$.54 & {} & {} & {} & {} & {} & {} & {}\\
\hline
\end{tabular}
\caption{Unlearning on CIFARSuper20. We show the results for forgetting a sub-class from a super class. The Original Model is trained on complete dataset. The Retrained Model is trained on retain dataset. We use a randomly initialized model as \textit{incompetent teacher} and the original model as \textit{competent teacher}. The ZRF score should increase on forget set after unlearning. The JS-Div: Jensen-Shannon Divergence, MR: Mushrooms, Acc.: Accuracy, Orig.: Original Model, Veh2: Vehicles2, Veg: Vegetables, ED: Electrical Devices, NS: Natural Scenes}
\label{CIFARSuper-20}
\end{table*}

\section{Experiments}
\label{sec_exp_results}
\textbf{Datasets used.} We evaluate our proposed method on image classification: CIFAR10~\cite{krizhevsky2009learning}, CIFAR100~\cite{krizhevsky2009learning}, epileptic seizure recognition~\cite{andrzejak2001indications}, and activity recognition~\cite{anguita2013public} datasets.\par

\textbf{Models used.} We use ResNet18, ResNet34, MobileNetv2, Vision Transformer, and AllCNN models for learning and unlearning in image classification tasks. We use a 3-layer DNN model for unlearning in epileptic seizure recognition. We use an LSTM model for unlearning in activity recognition task. All the experiments were performed on NVIDIA Tesla V100 (32 GB) with Intel Xeon processors. The experiments are implemented in PyTorch 1.5.0. The KL temperature is set to 1 for all the experiments.\par


\textbf{Evaluation Measures.} We use the following metrics for our analysis of the proposed unlearning method. \textit{1) Accuracy on forget \& retain set:} The accuracy of the unlearned model on $D_f$ and $D_r$ sets should be similar to the retrained model. \textit{2) Membership inference attack:} A membership inference attack is performed to check if any information about the forget samples is still remaining in the model. The attack probabilities should be lower on the forget set in the unlearned model. \textit{3) Activation distance:} This is an average of the L2-distance between the unlearned model and retrained model's predicted probabilities on the forget set. A lesser activation distance represents better unlearning. \textit{4) JS-Divergence:} JS-Divergence between the predictions of the unlearned and retrained model when coupled with activation distance gives a more complete picture on unlearning. Lesser the divergence, better the unlearning. \textit{5) ZRF score:} We introduce this metric to remove the dependence on the retrained model for evaluating the machine unlearning method.\par

\textbf{Baseline Models.} We use the unlearning method from~\cite{graves2021amnesiac} for a comparative analysis. This best fits our problem statement i.e., unlike most other methods, this method achieves unlearning in an already trained model without putting any constraints on the training procedure. We also use the retrained model for comparison. We perform two types of unlearning: (i) sample unlearning, and (ii) class unlearning. We present the experiments and analysis for each of them below.

\begin{figure}
\captionsetup{skip=0pt}
\centering
\begin{subfigure}{0.23\textwidth}
    \centering
    \includegraphics[width=1\textwidth]{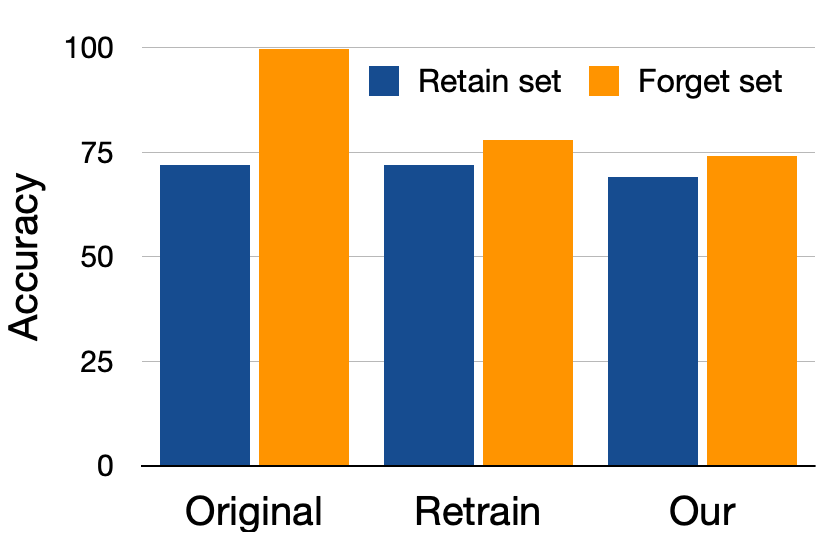}
    \caption{}
    \label{fig:eps50}
\end{subfigure}
\begin{subfigure}{0.23\textwidth}
    \centering
    \includegraphics[width=1\textwidth]{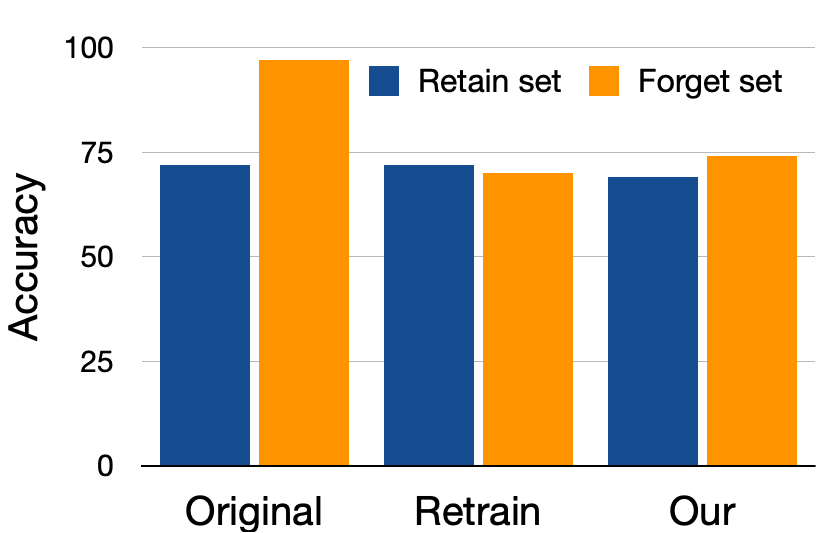}
    \caption{}
    \label{fig:eps100}
\end{subfigure}
\caption{Unlearning random samples (50 and 100 samples, respectively) from Epileptic Seizure Data.}
\label{fig:amnesiac}
\end{figure}
\subsection{Forget Acc. Vs Information Exposure Trade-off} \label{sec:forget-accuracy}
Machine unlearning of a specific class or cohort often leads to a decrease in accuracy or performance on forget set. Although, it is an expected result when the forget set is orthogonal to the retain set i.e., there are no samples in the retain set similar to the ones in the forget set. But this may not be true when retain set contains data points similar to the forget set samples. The accuracy may \textit{drop slightly}, \textit{may not drop at all}, or even \textit{increase in some cases}. An unlearning method should bring the forget set performance \textit{closer} to the gold (retrained) model instead of simply reducing it. If the performance of the unlearned model deviates a lot from the gold model, it could lead to \textit{Streisand effect}. This effect refers to unexpected behaviour of the model on forget samples which may leak information about that data. The leak could be in the form of being consistently \& maximally incorrect about only the forget samples, signalling that a deliberate effort was made to forget a selected set of samples. The aim should be to avoid this in order to ensure that the information about the forget set has been properly erased. For example, as mentioned earlier in the \textit{aeroplanes} example, when method unlearns \textit{Boeing aircraft}, if it is maximally wrong whenever it sees any \textit{Boeing aircraft} image and classifies it as \textit{sea}, \textit{animals}, \textit{mushroom}, etc., it will be suspicious. Other unseen aeroplanes will not be classified incorrectly so consistently.
That means it has not really erased the information of \textit{Boeing aircraft}. That information still exists which the model uses to be deliberately incorrect about the forget set.

\subsection{Sample Unlearning}
\textbf{CIFARSuper20.} The CIFAR100 is made up of 20 super classes i.e., there are different variants for each of these 20 classes. We merge all classes of the CIFAR100 into their super classes and convert it into a 20 class set named CIFARSuper20. Each class in CIFARSuper20 have 5 sub-classes, which are actual classes of CIFAR100. We conduct experiments on CIFARSuper20 by forgetting one sub-class from each super class. This setup makes unlearning more difficult than a regular scenario as we need to unlearn a sample/class without damaging the information of another sample/class that looks quite similar to it (for example, forget \textit{baby} from \textit{people}	super class consisting of \textit{baby, boy, girl, man, woman}).\par

We present unlearning results on ResNet18, ResNet34, and Vision Transformer. We use pretrained models to train/fine-tune for 5 epochs using Adam optimizer with a batch size of 256. The learning rate is 0.001 for final layer and 0.0001 for pretrained weight layers. A learning plateau with patience of 3 and reduce factor 0.5 is used. We conduct multiple runs (5 times) of our algorithm which didn't show any significant variation in performance (refer Table~\ref{CIFARSuper-20}). Therefore, we report the results of single run for all the models in this paper.\par

We unlearn various sub-classes from a super-class. We use 30\% of retain data and a single epoch of unlearning for all models. The learning rate of 0.0001 is used for unlearning. Table~\ref{CIFARSuper-20} shows unlearning results on ResNet18 and Vision Transformer~\cite{dosovitskiy2020image}. The evaluation is performed on all the metrics discussed earlier. It can be observed in Table~\ref{CIFARSuper-20} that performance of our method is very close to that of the retrained model. There is very low probability of membership inference attack on our unlearned model. The accuracy of our method on the forget and retain set when forgetting \textit{rocket} images from \textit{vehicles} is almost same as the retrained models. The membership inference attack probability on the model for samples from \textit{rocket} class drops to 0.002 from 0.982 after unlearning. The JS-Divergence between predictions of the retrained model and our model is 0.04 in the forget set. This implies the output distribution of unlearned model is very close to the retrained model. The ZRF score of our model becomes 0.99 from 0.87 after unlearning, thus indicating effective forgetting. Furthermore, Table~\ref{CIFARSuper20-Cheap} shows the unlearning results in ResNet34 with different types of teachers.\par

\begin{table*}[t]
\footnotesize
\centering
\begin{tabular}{c|c|c|cccc|cc}
\hline
 & \multirow{2}{*}{Forget Set} & \multirow{2}{*}{Accuracy} & \multirow{2}{*}{Original} & \multirow{2}{*}{Retrain} & \multirow{2}{*}{Amnesiac} & \multirow{2}{*}{\textbf{Our}} & \multicolumn{2}{c}{Activation Distance}\\
\cline{8-9}
& {} & {} & &  & {} & &  \multicolumn{1}{c}{Amnesiac} & \multicolumn{1}{c}{\textbf{Our}}\\
\hline
RN18+ & \multirow{2}{*}{Rocket} & $D_r$ $\uparrow$ & 85.78 & 85.79 & 84.79 & 85.05 & \multirow{2}{*}{0.70} & \multirow{2}{*}{\textbf{0.67}}\\
{C20} & {} & $D_f$ $\downarrow$ & 82 & 3 & 4 & 2 & {} & {}\\
\cline{2-9}
Dataset & \multirow{2}{*}{Baby} & $D_r$ $\uparrow$ & 85.67 & 85.73  & 84.64 & 85.18 & \multirow{2}{*}{\textbf{0.65}} & \multirow{2}{*}{\textbf{0.65}}\\
{} & {} & $D_f$ $\downarrow$ & 93 & 82 & 78 & 77 & {} & {}\\
\hline
DNN+ & {50} & $D_r$ $\uparrow$ & 71.91 & 72.26 & 76.04 & 82.69 & \multirow{2}{*}{0.77} & \multirow{2}{*}{\textbf{0.47}}\\
{Seizure} & {samples} & $D_f$ $\downarrow$ & 98 & 70 & 30 & 74 & {} & {} \\
\cline{2-9}
 Dataset & {100} & $D_r$ $\uparrow$ & 71.91 & 73.39 & 75.17 & 79.26 & \multirow{2}{*}{0.69} & \multirow{2}{*}{\textbf{0.42}}\\
& {samples} & $D_f$ $\downarrow$ & 96 & 73 & 40 & 70 & {} & {}\\
\hline
LSTM+ & {Person} & $D_r$ $\uparrow$ & 90.46 & 84.01 & 89.45 & 87.27 & \multirow{2}{*}{0.52} & \multirow{2}{*}{\textbf{0.14}}\\
{HAR} & {\#1} & $D_f$ $\downarrow$ & 100 & 99.13 & 53.03 & 94.24 & {} & {}\\
\cline{2-9}
 Dataset &  {Person} & $D_r$ $\uparrow$ & 90.46 & 89.68 & 87.04 & 86.6 & \multirow{2}{*}{0.49} & \multirow{2}{*}{\textbf{0.13}}\\
{} &  {\#3} & $D_f$ $\downarrow$ & 99.41 & 99.41 & 75.95 & 95.31 & {} & {}\\
\hline
\end{tabular}
\caption{Comparison of our method with the Amnesiac learning~\cite{graves2021amnesiac}}
\label{baseline-comparison} RN18: ResNet18, C20: CIFAR20, HAR: Human Action Recognition
\end{table*}

\textbf{Epileptic Seizure Detection.} The dataset consists of the status of seizure in medical patients. There are a total of 178 predictor variables and 5 classes. A 3-layer DNN is trained for classification. The model is trained for 50 epochs using Adam Optimizer with learning rate of 0.01 and plateau with patience 10 and reduce factor 0.1. We unlearn 50 and 100 randomly selected data points. The results are presented in Figure~\ref{fig:eps50} and Figure~\ref{fig:eps100}. We observe that the proposed method performance is close to the retrained model. 
The accuracy on the forget set indicates that we have indeed effectively unlearned the forget set as the forget accuracy is reduced from around 100\% to a generalized performance. For example, in case of forgetting 100 samples, the accuracy on the forget set drops from 90\% to 74\% in our method which is close the 70\% accuracy of the retrained model. 

\textbf{Human Activity Recognition.} This is a task of classifying the activity of a person using the readings collected from smartphone sensors that an individual is carrying with her. The observation were taken from 30 different persons. The dataset contains 6 different types of activities which can be classified using time-series data with sensors giving 9 readings at each time-step. An LSTM Model with 2 dense layers after each LSTM is trained to predict the activity. The model is trained for 50 epochs using Adam Optimizer with learning rate of 0.01 and plateau with patience 10 and reduce factor 0.1. Table \ref{baseline-comparison} contains the results of forgetting person 1 and person 3. Detailed results and effects of various parameters on unlearning are present in the supplementary material.\par

\begin{table}[t]
\centering
\resizebox{\columnwidth}{!}{
\begin{tabular}{c|c|c|cc|cc}
\hline
{} & \# $\mathcal{Y}_{f}$ & Acc. & Orig. & Retrain & UNSIR & \textbf{Our}\\
\hline
RN18+ & \multirow{2}{*}{1} & $D_r$ $\uparrow$ & 77.86 & 78.32 & 71.06 & \textbf{78.46} \\
 {C10} & {} & $D_f$ $\downarrow$ & 81.01 & 0 & \textbf{0} & 4.22 \\
\cline{2-7}
Dataset & \multirow{2}{*}{2} & $D_r$ $\uparrow$ & 78.00 & 79.15 & 73.61 & \textbf{79.22} \\
 {} & {} & $D_f$ $\downarrow$ & 78.65 & 0 & \textbf{0} & 9.94 \\
\hline
{RN18+} & \multirow{2}{*}{1} & $D_r$ $\uparrow$ & 78.68 & 78.37 & 75.36 & \textbf{77.00}\\
{Pre+C100} & {} & $D_f$ $\downarrow$ & 83.00 & 0 & \textbf{0} & \textbf{0}\\
\cline{2-7}
Dataset & \multirow{2}{*}{20} & $D_r$ $\uparrow$ & 77.84 & 79.73 & 75.38 & \textbf{77.78} \\
{} & {} & $D_f$ $\downarrow$ & 82.84 & 0 & \textbf{0} & 3.90 \\
\hline
AllCNN+ & \multirow{2}{*}{1} & $D_r$ $\uparrow$ & 82.64 & 85.90 & 73.90 & \textbf{81.74} \\
{C10} & {} & $D_f$ $\downarrow$ & 91.02 & 0 & \textbf{0} & 9.16 \\
\cline{2-7}
 Dataset & \multirow{2}{*}{2} & $D_r$ $\uparrow$ & 84.27 & 85.21 & \textbf{80.76} & 77.68 \\
 {} & {} & $D_f$ $\downarrow$ & 79.74 & 0 & \textbf{0} & 5.64 \\
\hline
MNv2+ & \multirow{2}{*}{1} & $D_r$ $\uparrow$ & 77.43 & 78 & 75.76 & \textbf{78.22} \\
{Pre+C100} & {} & $D_f$ $\downarrow$ & 90 & 0 & \textbf{0} & \textbf{0} \\
\cline{2-7}
Dataset& \multirow{2}{*}{20} & $D_r$ $\uparrow$ & 76.47 & 77 & 76.27 & \textbf{76.65} \\
{} & {} & $D_f$ $\downarrow$ & 81.70 & 0 & \textbf{0} & 13.65 \\
\hline
\end{tabular}
}
\caption{Class-level unlearning on CIFAR10 and CIFAR100. The results are compared with UNSIR~\cite{tarun2021fast}. C10: CIFAR10, C100: CIFAR100, RN18: ResNet18, MNv2: MobileNetv2, Pre: Pretrained}
\label{Class-UnLearning}
\end{table}

\textbf{Comparison with Amnesiac learning~\cite{graves2021amnesiac}.} 
We compare our result with Amnesiac learning which fine-tunes the model with random labels on forget samples. Table~\ref{baseline-comparison} shows the comparison between both the methods. We compare the \textit{activation distance} and \textit{accuracy} on the forget and retain set. A lower \textit{activation distance} indicates closeness to the retrained model. This subsequently indicates better unlearning and an accuracy closer to the retrained model is desired on forget and retain set. The \textit{activation distance} for our method is very low compared to Amnesiac method in most of the cases (refer Table~\ref{baseline-comparison}). Amnesiac method causes too much damage in the forget set of epileptic seizure and human activity recognition dataset, indicating Streisand effect. The accuracy in epileptic seizure dataset (forget set of 50 samples) is 98\% for the original model, 70\% for retrained model, 74\% for our method, and 30\% for amnesiac method. The Amnesiac method damages the performance on forget set by a huge margin. It reduces the forget set accuracy to 30\% which otherwise should be close to 70\%. It should also be noted that \textit{activation distance} from retrained model is 0.47 for our method and 0.77 for Amnesiac method. Amnesiac method is causing undesired effects and the generated model is very different from the retrained model. Our method, besides being more effective and robust, requires access to only a subset of retain data. We use only 30\% of retain data to obtain the results in our method. Our method is $\sim2\times$ faster than Amnesiac method, more effective even when limited data is available for use.

\subsection{Class Unlearning}
We also demonstrate full-class (single and multiple classes) unlearning capability of our method. We show results on CIFAR10 and CIFAR100 with ResNet18, AllCNN, and MobileNetv2 models. Class-level unlearning results are compared with an existing method with configuration as in~\cite{tarun2021fast}. The model update in our method is performed for 1 epoch using 30\% of the retain data. The learning rate at the time of unlearning is set to 0.001. Table~\ref{Class-UnLearning} gives a performance comparison between the proposed and the existing methods. The accuracy on the retain set in CIFAR10 single-class forgetting is 71.06\% for UNSIR, 78.32\% for the retrained model, and 78.46\% for our method. The results are quite similar in all three methods. The accuracy on forget set is zero in the retrained model and UNSIR but our method retains some accuracy on the forget set. This is because the method learns from a randomly initialized teacher which does random predictions and predicts each class with ~10\% probability and forget model learns the same. This in turn leads to better protection against the risk of privacy exposure. 

\begin{table}[t]
\centering
\resizebox{\columnwidth}{!}
{
\begin{tabular}{c|c|c|c|ccc}
\hline
\multirow{2}{*}{{$T_{d}$}} & Super & Sub & \multirow{2}{*}{Acc.} & Orig.  & Retrain & \textbf{Our}\\
 &  {Class} & {Class} &  & Model & Model &\textbf{Method}\\
\hline
\multirow{4}{*}{\rotatebox[origin=c]{90}{ResNet34}} & \multirow{2}{*}{Veh2} & \multirow{2}{*}{Rocket} & $D_r$ $\uparrow$ & 86.36 & 85.32 & 85.8\\
 &  &  & $D_f$ $\downarrow$ & 88 & 4 & 1\\
\cline{2-7}
 & \multirow{2}{*}{Veg} & \multirow{2}{*}{MR} & $D_r$ $\uparrow$ & 86.41 & 85.92 & 85.61\\
 &  &  & $D_f$ $\downarrow$ & 83 & 2 & 5\\
\hline
\multirow{4}{*}{\rotatebox[origin=c]{90}{ResNet18}} & \multirow{2}{*}{Veh2} & \multirow{2}{*}{Rocket} & $D_r$ $\uparrow$ &  86.36 & 85.32 & 85.86 \\
 &  &  & $D_f$ $\downarrow$ & 88 & 4 & 15\\
\cline{2-7}
 & \multirow{2}{*}{Veg} & \multirow{2}{*}{MR} & $D_r$ $\uparrow$ & 86.41 & 85.92 & 85.83 \\
 & &  & $D_f$ $\downarrow$ & 83 & 2 & 1\\
\hline
\multirow{4}{*}{\rotatebox[origin=c]{90}{Random}} & \multirow{2}{*}{Veh2} & \multirow{2}{*}{Rocket} & $D_r$ $\uparrow$ & 86.36 & 85.32 & 86.04 \\
 &  &  & $D_f$ $\downarrow$ & 88 & 4 & 5\\
\cline{2-7}
 & \multirow{2}{*}{Veg} & \multirow{2}{*}{MR} & $D_r$ $\uparrow$ & 86.41 & 85.92 & 83.37 \\
 &  &  & $D_f$ $\downarrow$ & 83 & 2 & 11\\
\hline
\end{tabular}
}
\caption{Forgetting sub-class from a super class on CIFARSuper20+ResNet34 using different types of cheaper incompetent teacher ($T_{d}$)}
\label{CIFARSuper20-Cheap}
\end{table}

\begin{table}[t]
\centering
\resizebox{\columnwidth}{!}{
\begin{tabular}{c|c|cccc|cc}
\hline
Super- &  \multirow{2}{*}{Acc.} & \multirow{2}{*}{Orig.}  & \multirow{2}{*}{Ret.} & \multirow{2}{*}{\textbf{(RI)}} & \multirow{2}{*}{\textbf{(P)}} & \multicolumn{2}{c}{JS-Div}\\
\cline{7-8}
 {Sub} & {} &  &  &  & & \textbf{(RI)} & \textbf{(P)}\\
\hline
Veh2- & $D_r$ $\uparrow$ & 85.8 & 85.8 & 85.1 & 85.1 & \multirow{2}{*}{0.04} & \multirow{2}{*}{0.02}\\
 {Rocket} & $D_f$ $\downarrow$ & 82 & 3 & 2 & 2 & {} & {}\\
\hline
Veg- & $D_r$ $\uparrow$ & 85.8 & 85.4 & 84.8 & 85.4 & \multirow{2}{*}{0.04} & \multirow{2}{*}{0.08}\\
{MR} & $D_f$ $\downarrow$ & 78 & 4 & 1 & 3 & {} & {}\\
\hline
\end{tabular}
}
\caption{Forgetting sub-class from a super class in CIFARSuper20. Our(RI): Using randomly initialized teacher, Our(P): Using a partially trained model (1 epoch on 50\% of retain data) as an incompetent teacher. Ret.: Retrain}
\label{CIFARSuper-20-proxy}
\end{table}

\subsection{Using a Simpler Model as an Incompetent Teacher} \label{sec:cheaper-teachers}
Our method does not place any constraints on the architecture of the incompetent teacher. Preferably the architecture should be kept same as the student for proper transfer of information. But it can be replaced with smaller models without significantly affecting the results. As the teacher is initialized with random weights, such behaviour can be obtained by a significantly smaller model, or even hard coded algorithms to generate random predictions. A cheaper teacher can make the unlearning process faster without compromising in the quality of unlearning. We replace the incompetent teacher with (i) a small randomly initialized Neural Network, and (ii) a random prediction generator. The random prediction generator first assigns equal probability to all classes and then a adds Gaussian noise to the predictions. The performance with these teachers is shown in Table~\ref{CIFARSuper20-Cheap}. 
With ResNet34 as teacher, the performance on $D_f$ (forget class: \textit{Rocket}) is 1\% while using the same model as teacher, 15\% while using ResNet18, and 5\% while using a random predictor as a teacher. Similarly, the performance on retain set while using ResNet34 as teacher is 85.8\%, 85.86\% while using ResNet18, and 86.04\% while using a random predictor as a teacher. There is a negligible change in performance when we use simpler models as teachers. Thus, it can be used to reduce the computational costs without much loss in the performance. 

\subsection{Using Partially Retrained Model as an Incompetent Teacher} \label{sec:proxy}
A partially trained (PT) model on a subset of retain data can be used as an incompetent teacher in the proposed framework. Similarly, smaller models trained on a small subset of the retain data can also serve as an incompetent teacher. We show the results of using PT models as incompetent teachers to induce forgetting. Similarly, we further investigate the effectiveness of PT teacher on CIFARSuper20 in Table~\ref{CIFARSuper-20-proxy}. The teacher is trained for 1 epoch on 50\% of the data. The accuracy on forget set \textit{Rocket} is 3\% for retrained model, 2\% for our method with PT teacher and 2\% for RI teacher. The accuracy on retain set for \textit{Rocket} is 85.79\% for retrained model, 85.07\% for our method with PT teacher and 85.05\% for our method with RI teacher. Besides, the JS-Divergence between retrained model \& RI model based unlearning is 0.04 and 0.02 in case of PT teacher model based unlearning. This shows that in addition to accuracy improvement, PT teacher based unlearning may also give output distribution more similar to the retrained model.



\subsection{Efficiency Analysis}
\label{sec: performance-comparison}
We compare the run-time comparison of the retrained model, the existing methods, and the proposed methods. The random weights based setup is $\sim70\times$ faster than retraining and more than $2\times$ faster than Amnesiac learning~\cite{graves2021amnesiac}. The method is faster when cheaper unlearning teachers are used. The proxy model based setup is about $20\times$ faster than retraining method. The ideal trade-off between efficiency and performance can be obtained by using smaller models partially trained on retain data but they come with the expense of additional training. This (partial training) further comes with a trade-off between computational cost and closeness of the model to the retrained model. The right amount of partial training should be decided. We observed that the cheap randomly initialized models are more efficient and generally perform well in most cases.\par

%% file: sec/5_conclusions.tex
\section{Conclusion}
We present a novel and general teacher-student framework for machine unlearning. A pair of competent and incompetent teachers is used to selectively transfer knowledge into the student network to obtain the unlearned model. Our work supports single \& multiple classes forgetting, sub-class forgetting and random samples forgetting. The effectiveness is evaluated in various application domains and modality of networks. We also introduce a new evaluation metric ZRF that is free from the need of having a retrained model for reference. This metric would be useful in real world scenarios where retrained models are not available or very expensive to obtain. Several possible efficient teachers are also explored to reduce the computational complexity. Future work could focus at the intersection of efficiency and privacy guarantees which may be in the form of either developing better evaluation measures or developing new class of unlearning techniques.



\section*{Acknowledgements}
This research is supported by the National Research Foundation, Singapore under its Strategic Capability Research Centres Funding Initiative. Any opinions,
findings and conclusions or recommendations expressed in
this material are those of the author(s) and do not reflect the views of National Research Foundation, Singapore.

%% file: sec/X_supplementary.tex
\newpage
\appendix
\onecolumn
\appendix
\begin{table*}[]
\centering
\caption{Forgetting the data (activity) of a single person in Human Activity Recognition Task. The results are presented for different number of epochs of unlearning with different size of subsets of retain data. The results are presented for an LSTM model. The JS-Divergence is computed between the retrained and our method}
\begin{tabular}{c|cc|ccc|ccc|c}
\hline
{No. of} & {\%} & \multirow{2}{*}{Accuracy} & Original  & Retrained & \textbf{Our} & \multicolumn{3}{c|}{ZRF} & \multirow{2}{*}{JS-Div}\\
\cline{7-9}
{epochs} & {of $D_r$} & {} & Model & Model &\textbf{Method} & \multicolumn{1}{c}{Original} & \multicolumn{1}{c}{Retrained} & \multicolumn{1}{c|}{\textbf{Ours}} & \\
\hline
\multirow{8}{*}{1} & \multirow{2}{*}{100\%} & $D_r$ $\uparrow$ & 90.46 & 84.01 & 88.43 & \multirow{2}{*}{0.57} & \multirow{2}{*}{0.72} & \multirow{2}{*}{0.79} & \multirow{2}{*}{0.03}\\
{} & {} & $D_f$ $\downarrow$ & 100 & 99.13 & 94.52 & {} & {} & {} & {}\\
\cline{2-10}
 & \multirow{2}{*}{50\%} & $D_r$ $\uparrow$ & 90.46 & 84.01 & 88.7 & \multirow{2}{*}{0.58} & \multirow{2}{*}{0.72} & \multirow{2}{*}{0.69} & \multirow{2}{*}{0.02}\\
{} & {} & $D_f$ $\downarrow$ & 100 & 99.13 & 97.69 & {} & {} & {} & {}\\
\cline{2-10}
 & \multirow{2}{*}{30\%} & $D_r$ $\uparrow$ & 90.46 & 84.01 & 88.77 & \multirow{2}{*}{0.58}  & \multirow{2}{*}{0.72} & \multirow{2}{*}{0.67} & \multirow{2}{*}{0.02}\\
{} & {} & $D_f$ $\downarrow$ & 100 & 99.13 & 98.27 & {} & {} & {} & {}\\
\cline{2-10}
 & \multirow{2}{*}{10\%} & $D_r$ $\uparrow$ & 90.46 & 84.01 & 90.19 & \multirow{2}{*}{0.58} & \multirow{2}{*}{0.72} & \multirow{2}{*}{0.62} & \multirow{2}{*}{0.01}\\
{} & {} & $D_f$ $\downarrow$ & 100 & 99.13 & 100 & {} & {} & {} & {}\\
\hline
\multirow{8}{*}{2} & \multirow{2}{*}{100\%} & $D_r$ $\uparrow$ & 90.46 & 84.01 & 88.22 & \multirow{2}{*}{0.57} & \multirow{2}{*}{0.71} & \multirow{2}{*}{0.79} & \multirow{2}{*}{0.02}\\
{} & {} & $D_f$ $\downarrow$ & 100 & 99.13 & 94.81 & {} & {} & {} & {}\\
\cline{2-10}
 & \multirow{2}{*}{50\%} & $D_r$ $\uparrow$ & 90.46 & 84.01 & 86.7 & \multirow{2}{*}{0.58}  & \multirow{2}{*}{0.72}  & \multirow{2}{*}{0.75} & \multirow{2}{*}{0.02}\\
{} & {} & $D_f$ $\downarrow$ & 100 & 99.13 & 96.83 & {} & {} & {} & {}\\
\cline{2-10}
 & \multirow{2}{*}{30\%} & $D_r$ $\uparrow$ & 90.46 & 84.01 & \textbf{87.75} & \multirow{2}{*}{0.58} & \multirow{2}{*}{\textbf{0.72}} & \multirow{2}{*}{\textbf{0.72}} & \multirow{2}{*}{0.03}\\
{} & {} & $D_f$ $\downarrow$ & 100 & 99.13  & \textbf{93.95} & {} & {} & {} & {}\\
\cline{2-10}
 & \multirow{2}{*}{10\%} & $D_r$ $\uparrow$ & 90.46 & 84.01 & 89.07 & \multirow{2}{*}{0.57} & \multirow{2}{*}{0.71} & \multirow{2}{*}{0.65} & \multirow{2}{*}{0.02}\\
{} & {} & $D_f$ $\downarrow$ & 100 & 99.13 & 99.71 & {} & {} & {} & {}\\
\hline
\multirow{8}{*}{5} & \multirow{2}{*}{100\%} & $D_r$ $\uparrow$ & 90.46 & 84.01 & 88.67 & \multirow{2}{*}{0.57} & \multirow{2}{*}{0.71} & \multirow{2}{*}{0.87} & \multirow{2}{*}{0.04}\\
{} & {} & $D_f$ $\downarrow$ & 100 & 99.13 & 89.91 & {} & {} & {} & {}\\
\cline{2-10}
 & \multirow{2}{*}{50\%} & $D_r$ $\uparrow$ & 90.46 & 84.01 & 86.90 & \multirow{2}{*}{0.56}& \multirow{2}{*}{0.71}& \multirow{2}{*}{0.84} & \multirow{2}{*}{0.04}\\
{} & {} & $D_f$ $\downarrow$ & 100 & 99.13 & 90.49 & {} & {} & {} & {}\\
\cline{2-10}
 & \multirow{2}{*}{30\%} & $D_r$ $\uparrow$ & 90.46 & 84.01 & 84.29 & \multirow{2}{*}{0.57} & \multirow{2}{*}{0.71} & \multirow{2}{*}{0.83} & \multirow{2}{*}{0.05}\\
{} & {} & $D_f$ $\downarrow$ & 100 & 99.13 & 91.07 & {} & {} & {} & {}\\
\cline{2-10}
 & \multirow{2}{*}{10\%} & $D_r$ $\uparrow$ & 90.46 & 84.01 & 85.71 & \multirow{2}{*}{0.57} & \multirow{2}{*}{0.71} & \multirow{2}{*}{0.75} & \multirow{2}{*}{0.04}\\
{} & {} & $D_f$ $\downarrow$ & 100 & 99.13 & 92.22 & {} & {} & {} & {}\\
\hline
\end{tabular}
\label{HAR-Ablation}
\end{table*}
\section{Additional Ablation Studies}
We study the effect of various hyper parameters in the unlearning method for the Human Activity Recognition dataset. We discuss the results achieved at different epochs of unlearning and at different learning rates below.

\subsection{Effect of Different Number of Epochs}
Without loss of generality, we unlearn the data related to user-1 from the activity dataset. Table~\ref{HAR-Ablation} shows how varying the size of the data used for retraining (\% of $D_r$) and the number of epochs for fine-tuning effects unlearning performance. The learning rate is fixed at 0.001. From Table~\ref{HAR-Ablation} we observe that using more data samples from $D_r$ result in better randomization in the outputs for $D_f$ in the unlearned model. For example, when only a single epoch is run and 100\% of $D_r$ is used, the ZRF score is 0.79. This goes down to 0.69 if only 50\% of $D_r$ is used. The ZRF score further declines to 0.67 and 0.62 if only $30\%$ and $10\%$ of $D_r$ is used, respectively. Increasing the number of epochs has a similar effect to that of increasing the number of samples from $D_r$. For example, with $10\%$ of $D_r$, the score ZRF is 0.62 after 1 epoch, 0.65 after 2 epochs and 0.75 after 5 epochs, respectively. The decision of selecting the number of epochs and \% of $D_r$ for retraining would depend on the degree of desired efficiency and the amount of $D_r$ available for use. 

\subsection{Effect of Different Learning Rates}
Table~\ref{HAR-LRAblation} shows the effect of using different learning rate during the brief retraining period. The number of epochs and the subset of $D_r$ are fixed at 2 and $30\%$, respectively. Increasing the learning rate increases the amount of randomization in the updated model. The ZRF score steadily goes down from 0.96 (learning rate 0.1) to 0.59 (learning rate 0.0001). 

Evidently, our method is highly flexible with respect to the amount of randomization desired by the model owner or the customer. Increasing the number of epochs, portion of the retain data or the learning rate increases the ZRF score. As explained in Section 4 in the main paper, we want the ZRF score to be as close to the retrained model as possible. The ZRF score closest to the retrained model is obtained with 30\% retain data, 2 epochs and a learning rate of 0.001. The JS-Divergence in this case between the retrained model and the model obtained by our method is 0.03. This indicates that the output distributions of the retrained model and the model obtained using our method are very similar.


\section{Sequential Unlearning} 
\begin{figure*}
    \centering
    \includegraphics[width=0.7\textwidth]{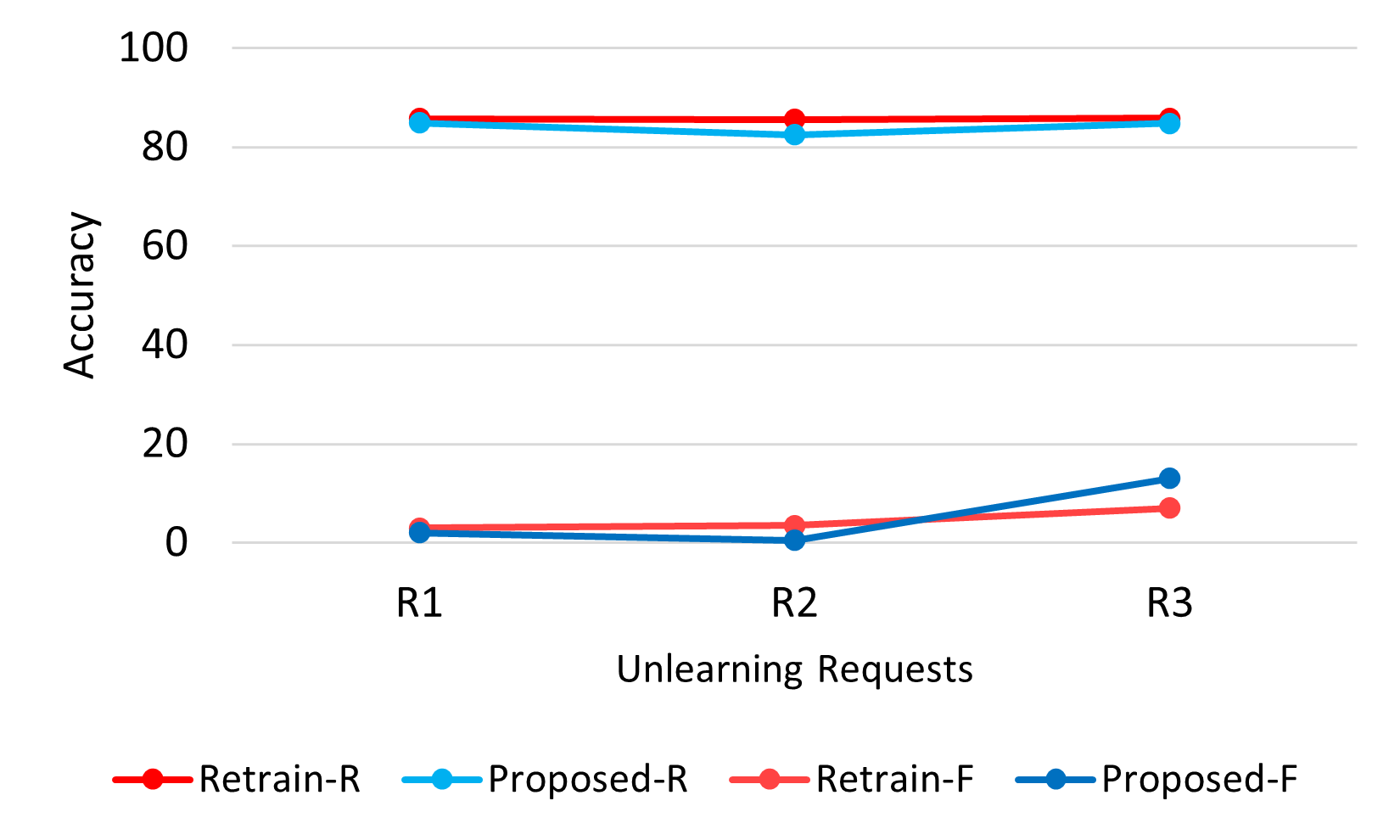}
    
    \caption{Sequential Unlearning Requests: Request 1 (R1), Request 2 (R2), Request 3 (R3) are the subsequent unlearning requests for subclasses Rocket, Mushroom and Lamp. We observe that the performance achieved by our method is very close to that of the model trained from scratch (retrained model) on both retain and forget set.}
    \label{fig:sequential}
\end{figure*}

We conduct experiments to simulate a real life scenario where unlearning may be requested repeatedly. Figure~\ref{fig:sequential} showcases the performance of our method in the following scenario: we forget sub-classes \textit{Rocket}, \textit{Mushrooms}, and \textit{Lamp} one after the other. We observe that our method performs really well here. Even after multiple requests of unlearning, there is hardly any degradation in performance which indicates the robustness of our method. 

\section{Partially Retrained Model as a Competent Teacher: Additional Experiments} \label{sec:proxy}
A partially trained (PT) model on a subset of retain data can be used as an incompetent teacher in the proposed framework. Similarly, smaller models trained on a small subset of the retain data can also serve as the incompetent teacher. We show the results of using PT models as incompetent teachers to induce forgetting. Table~\ref{Class-UnLearning-proxy} shows results of class unlearning using PT teacher. The teacher is trained for 2 epochs on $50\%$ of the retain data. The accuracy on forget set with ResNet18 is 0\% for retrained model, UNSIR, and our method with PT teacher. Whereas, it is 4.22\% in our method with randomly initialized (RI) teacher. The accuracy on retain set for ResNet18 is 78.32\% for retrained model, 78.6\% for our method with PT teacher, 78.46\%  for our method with RI teacher, and 71.06\% for UNSIR. The performance is closer to the retrained model when PT teacher is used instead of RI teacher.

\begin{table*}[]
\centering
\caption{Learning rate ablation on Human Activity Recognition task. The unlearning is done for 2 epochs with 30\% of the retain data.}
\begin{tabular}{cc|ccc|ccc|c}
\hline
{Learning} & \multirow{2}{*}{Accuracy} & Original  & Retrained & \textbf{Our} & \multicolumn{3}{c|}{ZRF} & \multirow{2}{*}{JS-Div}\\
\cline{6-8}
{Rate} & {} & Model & Model &\textbf{Method} & \multicolumn{1}{c}{Original} & \multicolumn{1}{c}{Retrained} & \multicolumn{1}{c|}{\textbf{Ours}} & \\
\hline
\multirow{2}{*}{0.1} & $D_r$ $\uparrow$ & 90.46 & 84.01 & 72 & \multirow{2}{*}{0.57} & \multirow{2}{*}{0.71} & \multirow{2}{*}{0.96} & \multirow{2}{*}{0.07}\\
{} & $D_f$ $\downarrow$ & 100 & 99.13 & 82.13 & {} & {} & {} & {}\\
\hline
\multirow{2}{*}{0.01} & $D_r$ $\uparrow$ & 90.46 & 84.01 & 86.19 & \multirow{2}{*}{0.57} & \multirow{2}{*}{0.72} & \multirow{2}{*}{0.93} & \multirow{2}{*}{0.07}\\
{} & $D_f$ $\downarrow$ & 100 & 99.13 & 88.76 & {} & {} & {} & {}\\
\hline
\multirow{2}{*}{0.001} & $D_r$ $\uparrow$ & 90.46 & 84.01 & 87.75 & \multirow{2}{*}{0.58}  & \multirow{2}{*}{\textbf{0.72}} & \multirow{2}{*}{\textbf{0.72}} & \multirow{2}{*}{0.03}\\
{} & $D_f$ $\downarrow$ & 100 & 99.13 & 93.95 & {} & {} & {} & {}\\
\hline
\multirow{2}{*}{0.0001} & $D_r$ $\uparrow$ & 90.46 & 84.01 & 90.4 & \multirow{2}{*}{0.57} & \multirow{2}{*}{0.71} & \multirow{2}{*}{0.59} & \multirow{2}{*}{\textbf{0.01}}\\
{} & $D_f$ $\downarrow$ & 100 & 99.13 & 100 & {} & {} & {} & {}\\
\hline
\end{tabular}
\label{HAR-LRAblation}
\end{table*}

\section{Efficiency Analysis: More Details}
\begin{figure*}[t]
\captionsetup{skip=0pt}
\centering
\begin{subfigure}{0.45\textwidth}
    \centering
    \includegraphics[width=1.23\textwidth]{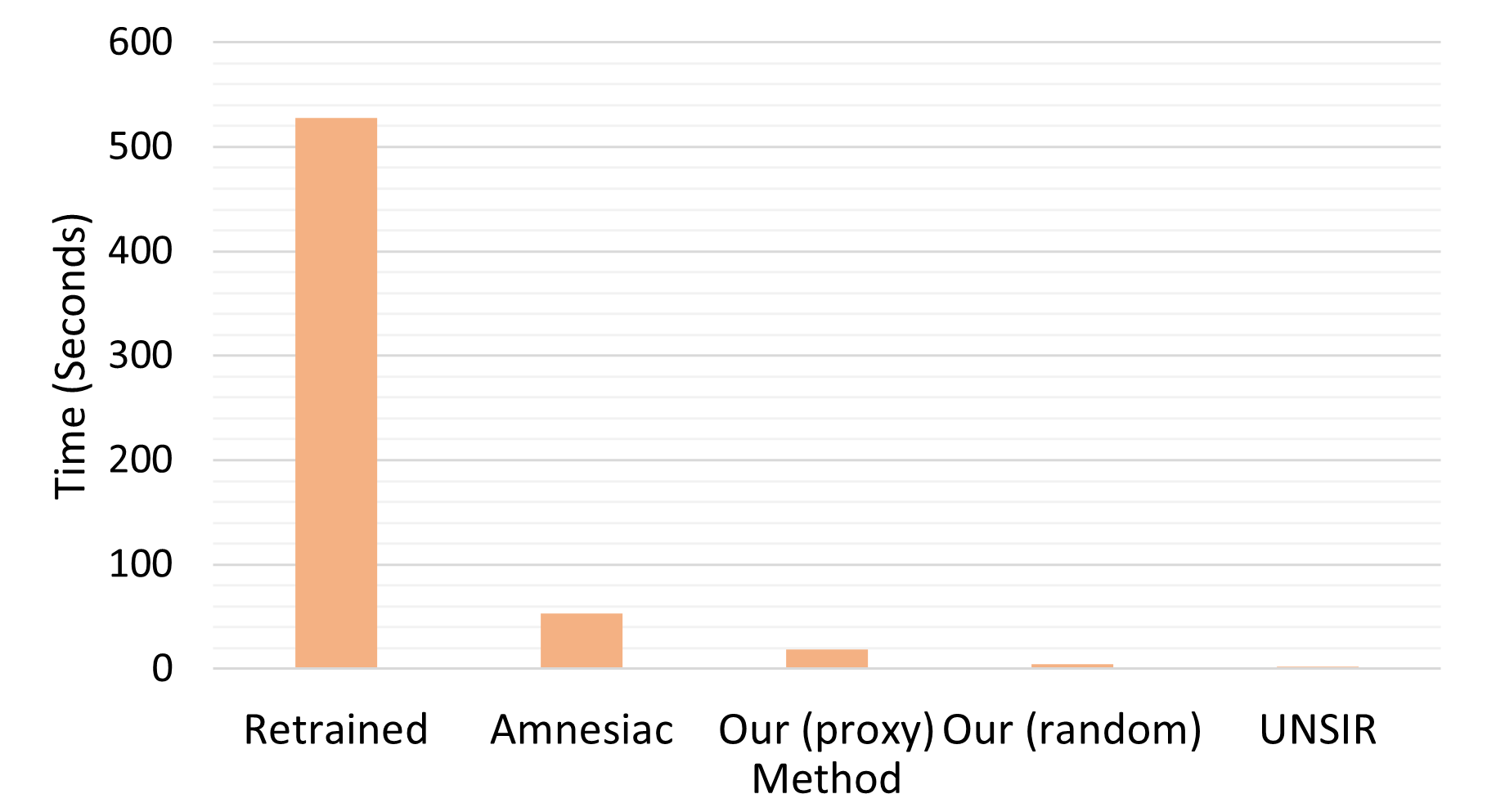}
    \label{fig:retrain-eff-comp}
\end{subfigure}
\begin{subfigure}{0.45\textwidth}
    \centering
    \includegraphics[width=1.1\textwidth]{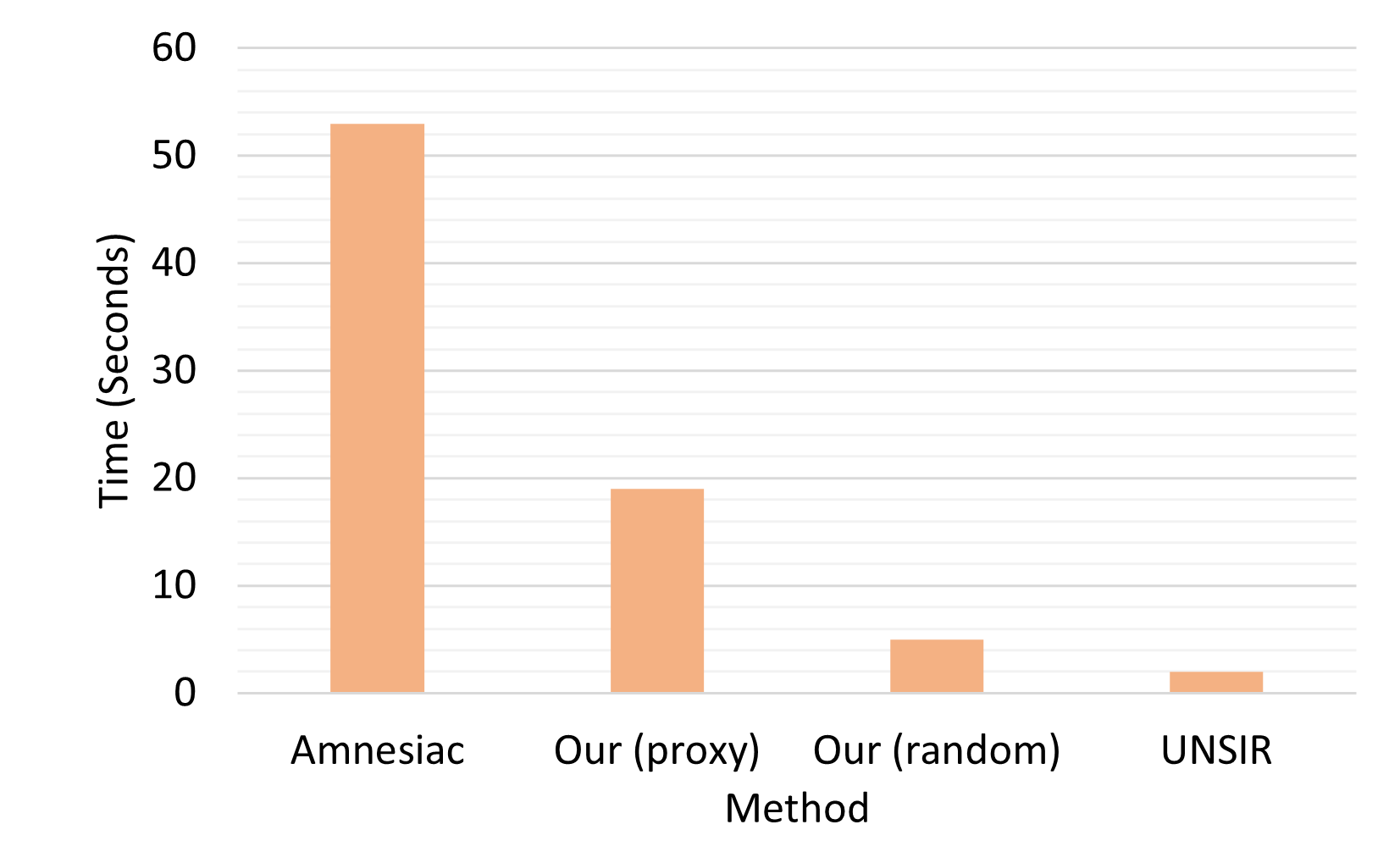}
    \label{fig:eff-comp}
\end{subfigure}
\caption{Efficiency comparison of various unlearning methods. We conduct the experiments for class unlearning in ResNet18 over CIFAR10.}
\label{fig:efficiency}
\end{figure*}

Figure~\ref{fig:efficiency} shows the total run-time comparison between the retrained model, the existing and the proposed methods. Our methods takes substantially less amount of time for unlearning in comparison to both the retrained method and the Amnesiac learning method~\cite{graves2021amnesiac}. Our method is only behind UNSIR~\cite{tarun2021fast} in run time efficiency. However, the UNSIR only supports class-level unlearning whereas, the proposed method supports both sample-level and class-level unlearning.

\begin{table*}[t]
\centering
\caption{Class-level unlearning for a single class in CIFAR10. Our(R): Using randomly initialized teacher, Our(P): Using partially trained model trained for 2 epochs on 50\% of retain data as incompetent teacher.}
\begin{tabular}{c|c|cc|ccc}
\hline
\multirow{2}{*}{Model}  & \multirow{2}{*}{Accuracy} & Original  & Retrained  & UNSIR & \multirow{2}{*}{\textbf{Ours(R)}}  & \multirow{2}{*}{\textbf{Ours(P)}} \\
{} & {} & Model & Model & \cite{tarun2021fast} & {} & {}\\
\hline
\multirow{2}{*}{ResNet18} & $D_r$ $\uparrow$ & 77.86 & 78.32 & 71.06 & 78.46 & \textbf{78.6}\\

{} & $D_f$ $\downarrow$ & 81.01 & 0 & \textbf{0} & 4.22 & \textbf{0}\\
\hline
\multirow{2}{*}{AllCNN}  & $D_r$ $\uparrow$ & 82.64 & 85.90 & 73.90 & 81.74 & \textbf{83.76}\\
{} & $D_f$ $\downarrow$ & 91.02 & 0 & \textbf{0} & 9.16 & 0.5\\
\hline
\end{tabular}
\label{Class-UnLearning-proxy}
\end{table*}

